\documentclass[10pt,twocolumn,letterpaper]{article}

\usepackage{wacv}              % To produce the CAMERA-READY version
\usepackage{graphicx}
\usepackage{amsmath}
\usepackage{amssymb}
\usepackage{booktabs}
\usepackage{colortbl}
\usepackage{xcolor}
\usepackage{balance}
\usepackage[accsupp]{axessibility}
\newcommand{\revise}[1]{{\color{black}#1}}

\definecolor{greenx}{RGB}{101,193,72}
\usepackage[pagebackref,breaklinks,colorlinks,citecolor=greenx]{hyperref}

\usepackage[capitalize]{cleveref}
\crefname{section}{Sec.}{Secs.}
\Crefname{section}{Section}{Sections}
\Crefname{table}{Table}{Tables}
\crefname{table}{Tab.}{Tabs.}

\def\wacvPaperID{866} % *** Enter the WACV Paper ID here
\def\confName{WACV}
\def\confYear{2024}

\usepackage{bm}

\begin{document}

%%%%%%%%% TITLE - PLEASE UPDATE
\title{Learning to Adapt CLIP for Few-Shot Monocular Depth Estimation}

\author{
Xueting Hu\textsuperscript{1} \quad 
Ce Zhang\textsuperscript{1} \quad  
Yi Zhang\textsuperscript{1} \quad 
Bowen Hai\textsuperscript{1} \quad  
Ke Yu\textsuperscript{1} \quad  
Zhihai He\textsuperscript{1,2}\thanks{Corresponding author.}\\
{\normalsize 
\textsuperscript{1}Department of Electronic and Electrical Engineering, Southern University of Science and Technology, Shenzhen, China}\\
{\normalsize 
\textsuperscript{2}Pengcheng Laboratory, Shenzhen, China}\\
{\tt\footnotesize \{huxt2022, zhangc2019, zhangyi2021, haibw2022, yuk2020\}@mail.sustech.edu.cn, hezh@sustech.edu.cn}
}
\maketitle

\begin{abstract}
Pre-trained Vision-Language Models (VLMs), such as CLIP, have shown enhanced performance across a range of tasks that involve the integration of visual and linguistic modalities. When CLIP is used for depth estimation tasks, the patches, divided from the input images, can be combined with a series of semantic descriptions of the depth information to obtain similarity results. The coarse estimation of depth is then achieved by weighting and summing the depth values, called depth bins, corresponding to the predefined semantic descriptions. The zero-shot approach circumvents the computational and time-intensive nature of traditional fully-supervised depth estimation methods. However, this method, utilizing fixed depth bins, may not effectively generalize as images from different scenes may exhibit distinct depth distributions. To address this challenge, we propose a few-shot-based method which learns to adapt the VLMs for monocular depth estimation to balance training costs and generalization capabilities. Specifically, it assigns different depth bins for different scenes, which can be selected by the model during inference. Additionally, we incorporate learnable prompts to preprocess the input text to convert the easily human-understood text into easily model-understood vectors and further enhance the performance. With only one image per scene for training, our extensive experiment results on the NYU V2 and KITTI dataset demonstrate that our method outperforms the previous state-of-the-art method by up to 10.6\% in terms of MARE\footnotemark.
\end{abstract}
\footnotetext{This work is supported by Center for Computational Science and Engineering at Southern University of Science and Technology.}

\begin{figure}[!th]
\centering
\includegraphics[width=\linewidth]{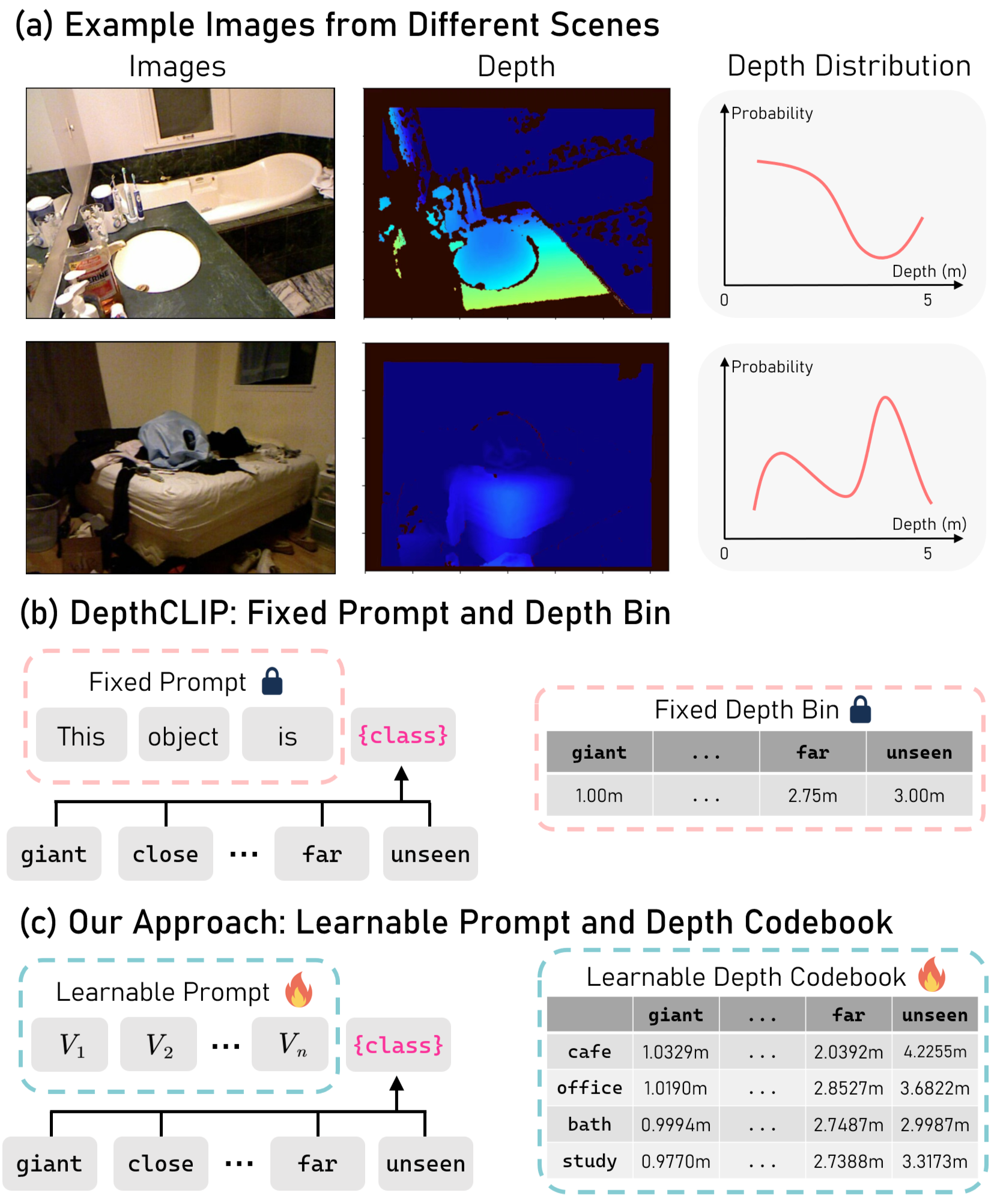}
\vspace{-18pt}
\caption{\textbf{Illustration of our motivation and major idea.} (a) Depth estimation and distributions of two example images from \texttt{bathroom} and \texttt{bedroom} scenes. (b) Fixed prompt and depth bin in DepthCLIP \cite{zhang2022can}. (c) Proposed learnable prompt and depth codebook in our approach.}
\vspace{-5pt}
\label{fig:intro}
\end{figure}

%%%%%%%%% BODY TEXT
\vspace{-3pt}
\section{Introduction}
\label{sec:intro}

Recently, attention has been drawn toward large-scale pre-trained Vision-Language Models (VLMs). 
VLMs are pre-trained on massive amounts of pairs of texts and images available on the Internet, allowing them to gain a deeper understanding of the connection between language and vision and learn more comprehensive visual representations \cite{radford2021learning,jia2021scaling}. 
Given these capabilities, VLMs can play a crucial role in processing multi-modal tasks.
Notably, the Contrastive Language-Image Pre-Training (CLIP)\cite{radford2021learning} model stands out for its remarkable image and text matching capability. 
During pre-training, CLIP encodes the input image and text using separate image and text encoders to obtain corresponding feature representations.
By measuring the cosine similarity between the image feature and text feature, CLIP identifies the image-text pair with the highest similarity as the positive sample and the rest as negative samples.
With a contrastive learning loss function, the CLIP model is trained on a dataset consisted of 400 million image-text pairs. Empirical evidence has demonstrated that such pre-trained models are highly effective in tasks such as few-shot image classification \cite{zhou2022learning,zhang2023unsupervised,zhang2023bdc}, transfer learning \cite{tang2023neuro,tang2023cross}, and image captioning \cite{li2021grounded,wang2022efficient}.

Monocular depth estimation, a key downstream task in computer vision, holds significant applications in autonomous driving, virtual reality and robotics.  
Traditional machine learning algorithms such as support vector machines (SVM) and random forests are commonly employed in this context \cite{roy2016monocular,fang2016fast}.
In contrast, deep learning-based depth estimation methods leverage deep neural networks to acquire intricate feature representations and perform depth prediction tasks. 
These methodologies typically exploit deep learning models like convolutional neural networks and autoencoders \cite{fu2018deep,tosi2019learning}.
Both machine learning-based and deep learning-based methods entail substantial data requirements for successful training, and are computationally expensive and time-consuming \cite{ming2021deep}.

Recently, a novel approach called DepthCLIP \cite{zhang2022can} proposes to perform zero-shot monocular depth estimation using the large-scale CLIP model. In this approach, the patches of the input image respond to a specific semantic distance token, which determine the contribution of this patch to various depth descriptions.
The depth descriptions, which are predefined by setting corresponding depth values manually, are then weighted and summed with the relevant depth values to yield the final depth estimation results. 
However, we recognize that images from different scenes have various depth distributions. As shown in Figure \ref{fig:intro} (a), the two images from different scenes have totally different depth distributions.
Therefore, pre-setting specific parameters, such as the depth values corresponding to the classes describing the depth, may lead to model performance that is strong in certain scenarios but suboptimal in others.

We observe that the key to achieving optimal performance for monocular depth estimation models lies in their ability to generalize effectively. 
To overcome the generalizability issues outlined previously, in this work, we design a few-shot learning approach. 
%这里写我们为什么选择用few shot 他能做什么
We endeavor to achieve a harmonious balance between the traditional approach, which can be time-consuming and labor-intensive, and the zero-shot method, which may result in limited generalization capabilities.
% Specifically, after the model has acquired a significant amount of data regarding a particular category, it can swiftly learn with just a small number of samples when presented with a brand new category.

\looseness=-1
Specifically, we propose a novel approach to adapt the CLIP model for few-shot monocular depth estimation.
Given its nature as a model architecture that integrates visual and textual elements, it is inherently advantageous to extract comprehensive and enhanced features from both visual and textual domains.
(1) From the visual perspective, an image can be examined globally or analyzed locally. A global view of a scene image allows for the extraction of its scene features. These scene features serve as evaluative criteria when determining the optimal depth bin selection.
Here, unlike previous methods which relied on artificially-set depth values corresponding to pre-defined depth descriptions, we design a depth codebook, which enables the model to identify the most appropriate depth value in accordance with  the current scene feature of the image during inference.
A local perspective of the image entails dividing it into patches, and for each patch, the image feature is extracted. All these extracted features play a crucial role in the subsequent stages of the process.
(2) From the textual perspective, we leverage a simple network to preprocess the input text by converting its content from easily human-understood text into easily model-understood vectors, enhancing the model's suitability for monocular depth estimation.
A comparison of our approach and previous state-of-the-art DepthCLIP \cite{zhang2022can} is shown in Figure \ref{fig:intro} (b) and (c).
Note that our approach is conducted with the few-shot setting. Specifically, we opt to utilize one image from each category for the purposes of training, during which several model parameters are adjusted and fine-tuned to yield optimal performance. With only one image per scene for training, our extensive experiment results on the NYU V2 dataset \cite{silberman2012indoor} demonstrate that our method outperforms the previous state-of-the-art method by up to 10.6\% in terms of MARE.

The major contributions of our paper are summarized as follows:
\begin{enumerate}
    \item We explore the monocular depth estimation task using vision-language models in a new few-shot setting, which aims to improve the depth estimation performance of pre-trained large-scale models by leveraging a small set of annotated samples.
    
    \item We recognize that images from different scenes have various depth distributions. To address this issue, we design learnable prompts and learnable depth codebooks to adapt the CLIP model for different scenes effectively.

    \item We evaluate our proposed method on NYU V2 dataset \cite{silberman2012indoor}. With only one image per scene for training, our method outperforms the previous state-of-the-art method by up to 10.6\% in terms of MARE, and can even be compared to fully-supervised methods.
\end{enumerate}

\section{Related Work}
\label{sec:related}
In this section, we review related works on vision-language models, prompt tuning methods for VLMs, and monocular depth estimation. 

\textbf{(1) Vision-language models.}
Large-scale pre-trained vision-language models (VLMs) have been developed to learn general visual representation under the supervision of natural languages  \cite{lei2015predicting,gomez2017self,sariyildiz2020learning,desai2021virtex,radford2021learning}. 
These models utilize extensive datasets to acquire representations that encompass the semantic comprehension of images and their accompanying textual descriptions.
For instance, CLIP \cite{radford2021learning} is derived through the application of contrastive learning on a dataset of 400 million curated image-text pairs, while ALIGN \cite{jia2021scaling} utilizes 1.8 billion noisy image-text pairs. 
Several other studies have been conducted along the direction of VLMs, like CoCa \cite{yu2022coca}, Flamingo \cite{alayrac2022flamingo}, and IDEA \cite{huang2022idea}.

In recent years, large-scale pre-trained VLMs have been applied to and shown great performances on various cross-modal alignment, zero-shot and few-shot image recognition tasks \cite{radford2021learning,gao2021clip,zhang2021vt,zhang2023cross}, and other visual tasks including image retrieval \cite{lu2019vilbert,duan2022multi}, visual grounding \cite{li2021grounded,yao2021cpt}, and visual question answering \cite{zhou2022unsupervised,duan2022multi,lei2021less}.

\textbf{(2) Prompt tuning methods for VLMs.}
The concept of ``prompt tuning" involves adapting pre-trained basic models for downstream tasks, especially in scenarios with limited or no prior training data. Initially introduced in the context of text input in language models, prompting aimed to enhance their output by providing specific instructions or examples as part of the input~\cite{radford2019language,brown2020language}. In subsequent research~\cite{schick2020exploiting, schick2020s}, the approach involved transforming the downstream task into a ``cloze" task using predefined patterns or templates, eliminating the need for task-specific classifiers. However, discovering the most effective patterns can be time-consuming, prompting recent studies to explore  soft prompts learned in a continuous manner~\cite{lester2021power,li2021prefix}.

\looseness=-1
Prompt tuning techniques for fine-tuning VLMs have focused on devising intricate prompts and incorporating adaptable context to extract task-relevant information from the encoded knowledge~\cite{zhou2022learning, zhou2022conditional}. In recent advancements, several notable  prompt tuning methods have demonstrated significant improvements \cite{lu2022prompt,wang2022learning,chen2023plot,jia2022vpt}.
CoOp~\cite{zhou2022learning}, a pioneering work in this field, optimizes prompt context by employing a series of learnable vectors, either in a unified or class-specific manner. Building upon CoOp, CoCoOp~\cite{zhou2022conditional} enhances the method by learning to generate vectors conditioned on individual image, effectively solving the problem of generalization to unseen classes.
%To tackle the issue of biases in prompts and handle varying visual representations, ProDA \cite{lu2022prompt} has been developed. ProDA focuses on learning low-bias prompts from a limited number of samples and is capable of capturing diverse prompt distributions. 
% DeFo~\cite{wang2022learning} introduces a prompt tuning approach that learns decomposed visual features with the assistance of feature-level textual prompts. PLOT~\cite{chen2023plot} focuses on learning multiple comprehensive prompts to describe the diverse characteristics of categories.
%Additionally, VPT~\cite{jia2022vpt} presents a prompt tuning method that performs full fine-tuning for large-scale transformer models in the vision domain. 

\textbf{(3) Monocular depth estimation.}
Monocular depth estimation is a crucial task in computer vision and finds numerous applications in autonomous driving, virtual reality, and robotics. Recent advancements in monocular depth estimation can be categorized into two main approaches: \textit{supervised} and \textit{unsupervised} methods.

\textit{Supervised methods} leverage ground truth depth maps during the training process to grasp semantic priors and extract semantic relationships \cite{casser2019depth,jackson2019style,la2022adversarial}. 
%In recent years, architectures like ResNet \cite{laina2016deeper}, U-Net \cite{wang2022u}, and Hourglass \cite{fu2018deep,liu2015learning} have been employed for depth estimation, while techniques like multi-task learning \cite{casser2019depth}, data augmentation \cite{jackson2019style}, and adversarial training \cite{la2022adversarial} have been used to improve the accuracy and robustness of the models.%
Recently, Li \textit{et al.} \cite{li2022depthformer} introduced Depthformer, which incorporates additional modules to enhance Transformer features through element-wise interaction. It models the affinity between Transformer and CNN features in a set-to-set translation manner, further improving depth estimation performance.
However, it would be costly and labor-intensive to gather fully-annotated data, which ultimately hinders the scalability of this approach.  

\textit{Unsupervised methods}, in contrast, typically try to pre-train the models to teach them to discover the semantic parts and objects within an image instead, and do not require additional annotated samples. 
Examples of unsupervised methods for monocular depth estimation include vid2depth \cite{mahjourian2018unsupervised}. This approach considers the 3D geometric structure of the scene and enforces consistency in estimated 3D point clouds and ego-motion on consecutive frames. 
Another work along this direction is from Zhang \textit{et al.} \cite{zhang2020unsupervised}, who also proposed a new unsupervised hybrid geometric-refined loss to explicitly explore the precise geometric relationship between the input color image and the predicted depth map. 

A recent notable approach called DepthCLIP \cite{zhang2022can} has gained significant attention. It leverages the text-image correlation capabilities of CLIP \cite{radford2021learning} to enable zero-shot depth prediction. In this work, we follow DepthCLIP \cite{zhang2022can} to further explore the potential of CLIP in depth estimation tasks.

%------------------------------------------------------------------------
\section{Method}
\label{sec:method}

\begin{figure*}[!th]
\centering
\includegraphics[width=\textwidth]{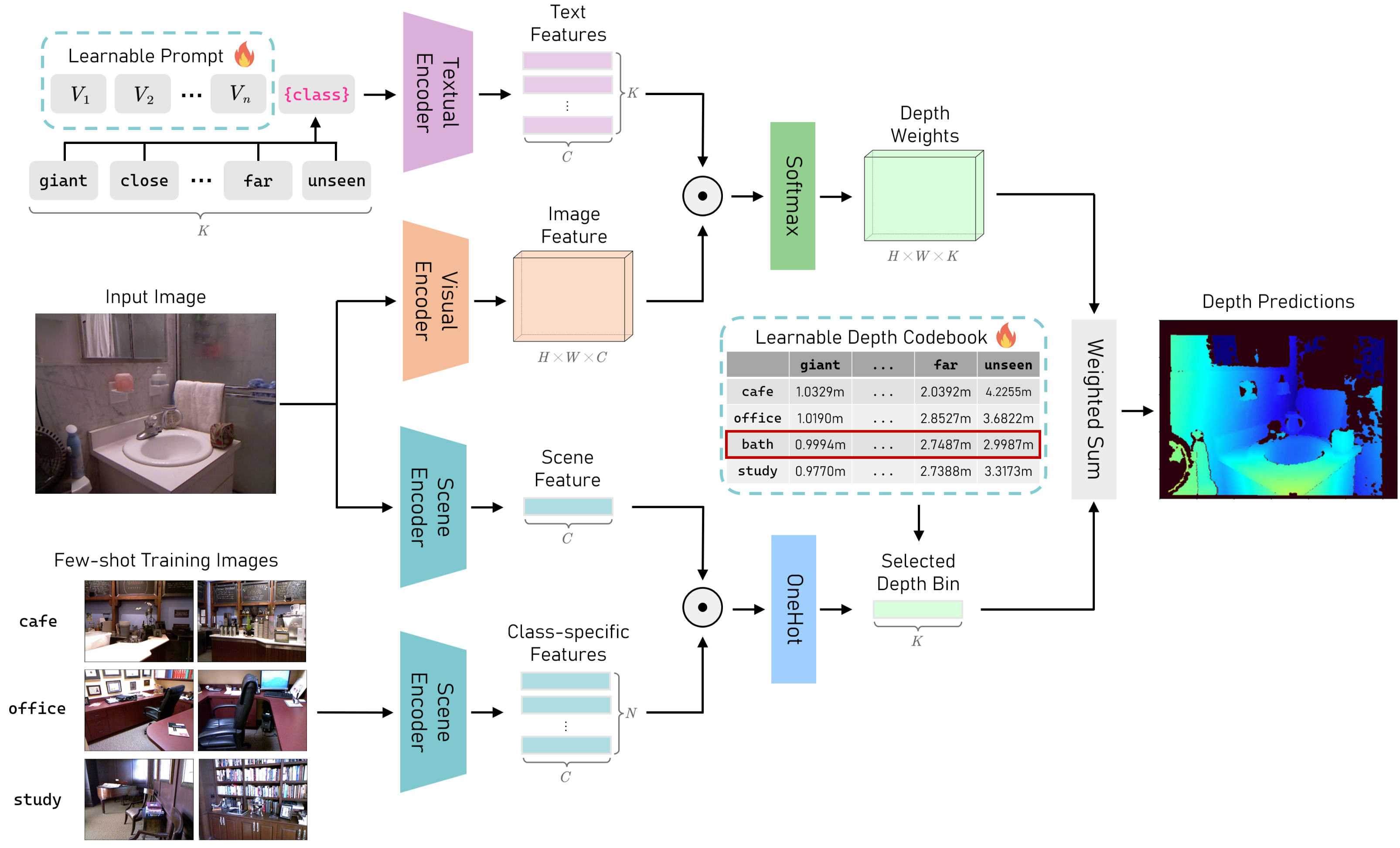}
\caption{\textbf{Overview of our proposed method.} We incorporate learnable prompts and learnable depth codebooks to adapt the CLIP model on few-shot samples for better depth estimation.  Three different encoders are included in this work. The textual encoder and scene encoder are directly from the original CLIP \cite{radford2021learning} model with ResNet-50 backbone. The visual encoder is the CLIP's ResNet-50 image encoder but without the final pooling layer. The fire icon means the parameters will be updated during few-shot training.
}
\vspace{-8pt}
\label{fig:overview}
\end{figure*}
In this section, we present our proposed method of learning to adapt CLIP for few-shot monocular depth estimation in detail. 

\subsection{Contrastive Language-Image Pre-Training}
The effectiveness of CLIP~\cite{radford2021learning} has been demonstrated in the multi-modal task of image-text matching. 
By leveraging contrastive learning methods, CLIP trains powerful text encoder and image encoder using a training dataset of 400 million text-image pairs.
Without the need for any training on new datasets, text and image features can be obtained separately by text encoder and image encoder, and the features are simply interacted with by dot product to obtain the image-text similarity, then the model can accurately select the text that best fits the input image from a number of texts based on the similarity.

% A CLIP model is denoted as $\{E_t, E_v\}$, where $E_t$ refers to the text encoder, and $E_v$ refers to the image encoder. In CLIP's original paper, the focus is on image classification, whereby a single test image $X_{test}$ belonging to a specific class $y$ is provided, where $X_{test}\in \mathbb{R} ^{C\times H\times W}$ and $y \in \mathbb{R} ^ K$ for a $K$-class classification problem. In the baseline zero-shot setting, each $y_i$ in the set $Y=\{y_1, y_2, \cdots, y_K\}$ is  concatenated with a pre-defined prompt such as $\rho =$ "a photo of," to create class-specific textual inputs, denoted as $\{\rho; y_i\}$. Text features, $\{t_1, t_2, \cdots, t_K\}$, are generated by the text encoder $E_t$, where $t_i = E_t({\rho; y_i})$. Subsequently, each text feature $t_i$ is combined with the image feature $v=E_v(X_{test})$ to compute a cosine similarity score, which can be used to predict the probability of $X_{test}$ belonging to class $y_i$.
% \begin{equation}
% \label{eq-sim}
%    sim\left( t_i,v \right)=\frac{t_i \cdot v}{\Vert t_i \Vert  \Vert v \Vert}.
% \end{equation}

% The prediction probability on $X_{test}$ can be denoted by 
% \begin{equation}
% \label{eq-clip}
%    p(y_i|X_{test})=\frac{\exp \left( sim\left( t_i,v \right) /\tau \right)}{\sum\nolimits_{j=1}^K{\exp \left( sim\left( t_j,v \right) /\tau \right)}}, 
% \end{equation}

% where $\tau$ is the temperature of the softmax function. We can easily identify the prediction $\hat{y}$ by:

% \begin{equation}
% \hat{y} = \operatornamewithlimits{argmax}_{y_i} p_{y_i}.
% \label{eq:pseudo_label}
% \end{equation}

To address the downstream task of depth estimation, we can leverage few-shot methods to extract information from limited data samples and conduct model fine-tuning. When constructing the model, we explore two main directions: 
In the text perspective, what kind of text is best understood by the model, how to better extract text features, and how to make the text features help the image task to the maximum extent.
In the image perspective, how to extract local and global information of an image and combine them with text to be applied to the downstream task of depth estimation.

\subsection{Method Overview}
In Figure \ref{fig:overview}, we present an overview of our proposed method for few-shot depth estimation.
The training process of our few-shot-based method requires one image from each scene category as a training sample. 
Preceding the training phase, we fetch all scene samples into the scene encoder to obtain a set of class-specific features. Simultaneously, we input all depth category descriptors into the text encoder to acquire a set of textual features.

\looseness=-1
During the training phase, the input image will be processed through the visual encoder and scene encoder, yielding image feature $\mathbf{F}_v$ through patch stitching and scene feature $\mathbf{F}_s$ through encoding the entire image. Subsequently, the cosine similarity is computed between the image feature and the set of textual features to estimate the depth category for each patch. The resulting similarity scores are then normalized to generate the depth estimation weight.
Likewise, the scene feature and the set of scene features undergo cosine similarity calculations, enabling the identification of the most similar scene within the scene set for the current input image. This identified scene is then utilized to select the corresponding depth bin from the depth codebook using the one-hot coding approach.
The depth estimation weight is then suitably weighted and merged with the depth bin to obtain the final depth values on a patch-by-patch basis. The depth of all pixels within each patch is equal to the depth of that patch.

\subsection{Learnable Prompt}

The prompt text for manually designed inputs in CLIP is characterized by textual features, while in DepthCLIP\cite{zhang2022can}, the textual input takes the form of "This object is \texttt{[Depth CLASS]}."
We artificially classify the continuous variable of depth into \texttt{[Depth CLASS]} and map it to discrete values.
 Specifically, the \texttt{[Depth CLASS]} includes ``giant", ``extremely close", ``close", ``not in distance", ``a little remote", ``far", and ``unseen" with corresponding values that can be mapped onto numerical measurements. For example, ``unseen" may represent a distance of 5 m.

 Although text prompts designed for human comprehension conform to English semantic and grammatical rules, they are often not intuitive for the model. 
 Consequently, in our method, we employed few-shot training techniques to convert these human-friendly words (``This", ``object", ``is") into computationally-friendly vectors, denoted as $\mathbf{V}=[\mathbf{v}_1, \mathbf{v}_2, ..., \mathbf{v}_n]$, where $n$ is a hyperparameter, which represents the number of vectors after converting words into vectors, the number of vectors does not need to match that of words. $\mathbf{v}_i$ is a row vector, representing a word token.  
 To ensure the model's sensibility toward various depth classes in the textual input, we adopt a unified text encoder, which takes the form of ``\texttt{[Vector]} + \texttt{[Class]}". Suppose we have $K$ depth classes and those $K$ classes are represented by $K$ tokens $\mathbf{m}_1,\cdots,\mathbf{m}_K$. This approach facilitates model comprehension and processing of textual features in these prompts.
\begin{equation}
\begin{aligned}
    \mathbf{T} = 
    \begin{bmatrix}
        \textbf{V} & \textbf{m}_1 \\
        \textbf{V} & \textbf{m}_2 \\
        \vdots & \vdots \\
        \textbf{V} & \textbf{m}_K
    \end{bmatrix}.
\end{aligned}
    % T = [Vec] [Depth CLASS].
    \label{Input Text}
\end{equation}

All these prompts can be transformed into text features $\mathbf{F}^{\star}_{t,i}$ via a text encoder:
\begin{equation}
    \mathbf{F}^{\star}_{t,i} = \mathbf{\Phi}_t(\mathbf{T}_i),
    \label{Text Encoder}
\end{equation}
where text features $\mathbf{F}^{\star}_{t,i} \in \mathbb{R}^{1 \times C}$, $i=1,\cdots,K$, $\mathbf{\Phi}_t$ represents the text encoder, it is directly from CLIP model, and $T_i$ is the $i$-th row of $\mathbf{T}$, $C$ is the length of the feature. 

In the training stage, the prompt vectors can be updated by the root-mean-square error of the depth estimation result with the ground truth, given by
\begin{equation}
    \mathcal{L}=\sqrt{\frac{1}{hw}\sum^h_{i=1}\sum^w_{j=1}(\hat{d}-d_{g.t.})^2},
    \label{loss}
\end{equation}
where $h$,$w$ are given by the original size of the input image, $\hat{d}$ and $d_{g.t.}$ denote predicted depth and ground truth depth value for each pixel, respectively.

\subsection{Learnable Depth Codebook}
The depth codebook $\mathbf{\Theta}$ is the key to convert this discrete value of classification results into depth values.
Our sample is composed by taking one image from each class. Accordingly, each scene is given a depth bin in our codebook.
%We get one sample from every scene class
%It gives a scene depth set to each scene [SceneCLASS], and each scene depth set is given a different depth value to the depth class [DEPTH\_CLASS]. 

For the scene images $\mathbf{S}_j$ used for the few-shot, they are first fed into the scene encoder $\mathbf{\Phi}_s$ to obtain the class-specific features 
\begin{equation}
    \mathbf{F}^{\star}_{s,j} = \mathbf{\Phi}_s(\mathbf{S}_j).
\end{equation}
The scene encoder is directly from the CLIP's ResNet-50 visual backbone. The class-specific features $\mathbf{F}^{\star}_{s,j} \in \mathbb{R}^{1 \times C}$, $j = 1, ... , N$, where $N$ is the number of scenes in the depthcodebook $\mathbf{\Theta}$. 

When the image $\mathbf{I}$ is input in, $\mathbf{I}$ went through the scene encoder to get the input scene features  $\mathbf{F}_s$.
\begin{equation}
    \mathbf{F}_s=\mathbf{\Phi}_s(\mathbf{I}),
    \label{Scene Encoder}
\end{equation}
where $\mathbf{F}_s\in\mathbb{R}^{1 \times C}$.

Consequently, we compute the similarity between the scene features of the input image and the class-specific features, facilitating the identification of the most appropriate depth bin within the codebook $\mathbf{\Theta}$ through a one-hot encoding technique. The similarity of the input image $\mathbf{I}$ to the scene image $s^s_j$ is calculated by
\begin{equation}
    s^s_j=\frac{\mathbf{F}_s {\mathbf{F}^{\star}_{s,j}}^{\top}}{\parallel \mathbf{F}_s\parallel \parallel \mathbf{F}^{\star}_{s,j}\parallel}.
    \label{similarity}
\end{equation}
The evaluated scene class $j^{\star}$ of the image $\mathbf{I}$ is expressed as
\begin{equation}
    j^{\star} = \underset{j}{\arg\max}\;s^s_j.
\end{equation}
Then the selected depth bin is the $j^{\star}$-th row of $\mathbf{\Theta}$:
\begin{equation}
    \bm{\theta}^{\star}=\mathbf{\Theta}_{j^{\star}}.
    \label{Depth Bin}
\end{equation}
The codebook $\mathbf{\Theta}$ is assigned an initial value and subsequently updated throughout the training process by the RMSE loss given in Equation (\ref{loss}).

\subsection{Depth Prediction}
\looseness=-1
Upon completing the tasks related to the image scene analysis and input text pre-processing, we proceed to promote the interaction between the input text and the input image.
For the input image $\mathbf{I}$, we  feed it into the image encoder without the pooling layer to obtain the image feature $\mathbf{F}_v$:
\begin{equation}
    \mathbf{F}_v=\mathbf{\Phi}_v(\mathbf{I}),
    \label{Image Encoder}
\end{equation}
where $\mathbf{F}_v\in\mathbb{R}^{H \times W \times C}$ and $\mathbf{\Phi}_v$ denotes the visual encoder, $HW$ is the number of image patches and $C$ is the feature space dimension.
Here, the visual encoder is slightly different from the scene encoder. It is also from the ResNet50-CLIP but without the final pooling layer.

Then, we calculate the cosine similarity between $\mathbf{F}^{\star}_{t,i}$ and $\mathbf{F}_v$ to get its similarity score
\begin{equation}
    s^v_i=\frac{\mathbf{F}^{\star}_{t,i} \mathbf{F}_v^{\top}}{\parallel \mathbf{F}^{\star}_{t,i}\parallel \parallel \mathbf{F}_v\parallel}.
    \label{similarity score}
\end{equation}

The similarity score vector $\mathbf{s}^v=[s^v_1,\cdots,s^v_K]$ is softmaxed to get the depth weight $\mathbf{d}$:
\begin{equation}
    \mathbf{d}=\mathrm{Softmax}(\mathbf{s}^v).
    \label{S Weight}
\end{equation}

Depth weight $\mathbf{d}\in\mathbb{R}^{H \times W \times K}$, $H \times W$ is the number of image patches and $K$ is the number of depth description class. 
The dimensionality of such a tensor means that for each of all $H \times W$ patches, its probability distribution across the $K$ depth categories is provided.

At this juncture, we have optimized and consolidated the accessible textual data, integrated and leveraged the global and local image information, and derived the depth weights and depth boxes. 
Ultimately, a straightforward weighted summation suffices to acquire the depth estimation $d$ for each patch
\begin{equation}
    \hat{d}=\bm{\theta}^{\star}\mathbf{d}^{\top}.
    \label{Depth}
\end{equation}

The depth estimation result obtained here is based on patches, and the depth of each pixel in each patch is equal to the value of the depth estimation result of the patch.

\section{Experiments}
\label{sec:experiment}
In this section, we present performance comparisons with state-of-the-art methods on monocular depth estimation, and ablation studies to demonstrate the effectiveness of our proposed method.

\subsection{Datasets}
\label{subsec:Dataset}
\textbf{NYU V2}~\cite{silberman2012indoor} is a dataset for indoor scene depth estimation.
It consists of 1,449 RGB images and corresponding depth maps for indoor scenes. 
The resolution of each image is 480 $\times$ 640, and the depth range for each pixel is 0-10m. 
The training set covers 36,253 images from 249 scenes, while the testing set includes 654 images from the remaining 215 scenes. 
To remove frames, all samples are cropped to a resolution of 416 $\times$ 512 pixels. 

\revise{
\textbf{KITTI}~\cite{geiger2013vision} is a large scale outdoor dataset with a resolution of 375 $\times$ 1242 pixels. It consists of RGB and depth image pairs captured by cameras and depth sensors in an autonomous driving car.
For a fair comparison, we adopt the split strategy of ASTransformer~\cite{chang2021transformer}. The training set is composed of 23,488 from 32 scenes while the testing set contains 697 images from the remaining 29 scenes. The training and 
testing samples are cropped to a resolution of 352 $\times$ 704 pixels.
}

\subsection{Implementation Details}
\label{subsec:Implementation Details}
We used PyTorch to build the model framework. 
Our method is built upon the pre-trained ResNet-50 CLIP model.
For the learnable prompts, we directly initialize them randomly. 
We selected one image per scene category to participate in the few-shot training. 
For the NYU V2~\cite{silberman2012indoor} dataset, we set  the initial dimension of the depth codebook as 27 $\times$ 7. 
Specifically, there are 27 scene categories to choose from, and the depth of each scene is manually classified into 7 categories: [`giant', `extremely close', `close', `not in distance', `a little remote', `far', `unseen'], with the same initial depth values assigned to these descriptive words: [1.00, 1.50, 2.00, 2.25, 2.50, 2.75, 3.00]. 
A depth bin contains a set of depth values, and for each scene class, we construct a class-dependent depth bin. Our depth codebook contains 27 class-dependent depth bins, which will be updated in subsequent training. 

In our experiments, the learning rate is set to 0.5 for the prompt training and 0.01 for the depth codebook training, a decay factor of $1\times 10^{-5}$  is used in both cases to prevent overfitting. 
After training for 200 epochs, we obtained the current results. All the experiments are conducted on a single NVIDIA RTX 3090 GPU.

% We used PyTorch to build the model framework. 
% The image and text encoders are from the pre-trained CLIP model, with a Transformer as the backbone model for the text branch and a ResNet-50 as the backbone model for the image branch. 
% For the text input prompts, we directly initialize vectors randomly. 
% We selected one image per scene category to participate in the few-shot training. 
% For the NYU V2~\cite{silberman2012indoor} dataset, we set  the initial dimension of the depth codebook as $27 \times 7$. 
% Specifically, there are 27 scene categories to choose from, and the depth of each scene is manually classified into 7 categories: [`giant', `extremely close', `close', `not in distance', `a little remote', `far', `unseen'], with the same initial depth values assigned to these descriptive words: [1.00, 1.50, 2.00, 2.25, 2.50, 2.75, 3.00]. 
% A depth bin contains a set of depth values, and for each scene class, we construct a class-dependent depth bin. Our depth codebook contains 27 class-dependent depth bins, which will be updated in subsequent training. In our experiments.

% The learning rate is set to $0.5$ during the prompt training and $0.01$ during the depth codebook training with a decay factor of $1\times 10^{-5}$ used in both cases to prevent overfitting. 
% After training for 200 epochs, we obtained the current results. All the experiments are conducted on a single NVIDIA RTX 3090 GPU.

\begin{table*}[t]
\vspace{0.3cm}
\centering
\resizebox{\linewidth}{!}{
	\begin{tabular}{l|cc|ccc|ccc}
	\toprule
		Method & Pre-training & Supervision & $\delta<1.25 \uparrow$ & $\delta<1.25^{2} \uparrow$ & $\delta<1.25^{3} \uparrow$ & MARE $\downarrow$ & AELS $\downarrow$ & RMSE $\downarrow$ \\ \midrule
		 Make3D~\cite{saxena2008make3d} &- &full  &0.447 &0.745 &0.897 &0.349 & - &1.214\\
		 DORN~\cite{fu2018deep}&- &full &0.828 &0.965 &0.992 &0.115 &0.051 &0.509\\
		 ASTransformer~\cite{chang2021transformer} &- &full &0.902 &0.985 &0.997  &0.103 &0.044 &0.374\\
		 DepthFormer~\cite{li2022depthformer}  &- &full &0.921	&0.989	&\textbf{0.998} &0.096	&0.041 &0.339\\
		 RPSF~\cite{mel2022end}  &- &full &\textbf{0.952} &\textbf{0.989} &0.997	&\textbf{0.072}	&\textbf{0.029}		&\textbf{0.267} \\
          \midrule
          LORN~\cite{li2019few} 	&ImageNet-1k~\cite{deng2009imagenet} &few-shot$^\dagger$ &0.703 &0.923 &0.979 &1.008 &0.222 &- \\ 
		 \midrule
		 Lower Bound &- &- &0.140 &0.297 &0.471 &1.327 &0.323 &2.934\\
		 vid2depth~\cite{mahjourian2018unsupervised}  &KITTI video~\cite{geiger2013vision} &0-shot &0.268	&0.507	&0.695 &0.572 &- &1.637\\
 	Zhang \textit{et al.}~\cite{zhang2020unsupervised}  &KITTI video~\cite{geiger2013vision} &0-shot &0.350	&0.617	&0.799 	&0.513 &0.529 &1.457\\ 

	    DepthCLIP~\cite{zhang2022can} 	&CLIP~\cite{radford2021learning} &0-shot &0.394 &0.683 &0.851 &0.388 &0.156 &1.167 \\
     \rowcolor{gray!20}
     \textbf{Ours} 	&CLIP~\cite{radford2021learning} &1-shot &\textbf{0.428} &\textbf{0.732} &\textbf{0.898} &\textbf{0.347} &\textbf{0.140} &\textbf{1.049}\\ 
	\bottomrule
	\end{tabular}
}
\vspace{-0.09cm}
\caption{\textbf{Performance comparisons of our proposed method and previous state-of-the-art methods on the NYU V2 dataset~\cite{silberman2012indoor}.} Lower bound is obtained by randomly making predication for each pixel within depth range 0-10m. We report the results obtained by previous state-of-the-art on fully-supervised and zero/few-shot settings. \revise{$^\dagger$LORN uses 200 images and 2,500 partial images for training.} Note that our method uses only one image per scene for better estimation. }
\label{tab:nyu_results}
\vspace{-10pt}
\end{table*}

\begin{table*}[t]
\vspace{0.3cm}
\centering
\resizebox{\linewidth}{!}{
	\begin{tabular}{l|cc|ccc|cccc}
	\toprule
		Method & Pre-training & Supervision & $\delta<1.25 \uparrow$ & $\delta<1.25^{2} \uparrow$ & $\delta<1.25^{3} \uparrow$ & MARE $\downarrow$ & MSRE $\downarrow$ & RMSE $\downarrow$ & AELS $\downarrow$ \\ \midrule
		 DORN~\cite{fu2018deep}&- &full &0.932 &0.984 &0.994 &0.072 &0.307 &2.727 &0.120\\
		 ASTransformer~\cite{chang2021transformer} &- &full &0.963 &0.995 &0.999  &0.103 &- &-  &0.374\\
		 DepthFormer~\cite{li2022depthformer}  &- &full &\textbf{0.975}	&\textbf{0.997}	&\textbf{0.999} &\textbf{0.052}	&\textbf{0.158} &\textbf{2.143} &\textbf{0.079}\\
		 \midrule
		
	    DepthCLIP~\cite{zhang2022can} 	&CLIP~\cite{radford2021learning} &0-shot &0.281 &0.531 &0.696 &0.473 &6.007 &12.958 &0.680 \\ \rowcolor{gray!20}
     \textbf{Ours} 	&CLIP~\cite{radford2021learning} &1-shot &\textbf{0.312} &\textbf{0.569} &\textbf{0.739} &\textbf{0.384} &\textbf{4.661} &\textbf{12.290} &\textbf{0.632}\\ 
	\bottomrule
	\end{tabular}
}
\vspace{-0.09cm}
\caption{\revise{\textbf{Performance comparisons of our proposed method and previous state-of-the-art methods on the KITTI dataset~\cite{geiger2013vision}.} The depth range on this task is 0-80m. We manually fine-tune the depth bin values for DepthCLIP~\cite{zhang2022can} to obtain these results. Note that in our method, the depth bins are learnable, so we only need to initialize the depth bins randomly. We only use 27 images to train our model.}}
\label{tab:kitti_results}
\vspace{-7pt}
\end{table*}

\subsection{Evaluation Metrics}
\label{subsec:Evaluation Metrics}
We evaluated the effectiveness of the method using four common metrics, which are divided into accuracy metrics and error metrics. 
The error metrics include the mean absolute relative error (MARE), \revise{mean squared relative error (MSRE)}, absolute error in log space (AELS), and root mean square error (RMSE). 
% The mean absolute relative error is the average of the depth estimation errors, reflecting the average proportion of measurement depth errors to true depth values. The root mean square error is the standard deviation of the depth estimation errors. The absolute error in log space calculates the average absolute difference between the true depth and predicted depth after taking the logarithm of both. The absolute error in log space can prevent underfitting and is more robust in processing depth images with a wider measurement range. 
These error metrics we used can be computed by
\begin{equation}
    \mathrm{MARE} = \frac{1}{N} \sum_{i=1}^{N} \frac{(|d_i-\hat{d_i}|)^2}{d_i},
    \label{MARE}
\end{equation}
\revise{
\begin{equation}
    \mathrm{MSRE} = \frac{1}{N} \sum_{i=1}^{N} \frac{|d_i-\hat{d_i}|}{d_i},
    \label{MSRE}
\end{equation}
}
\begin{equation}
    \mathrm{AELS} = \frac{1}{N} \sum_{i=1}^{N} |\log(d_i) - \log(\hat{d_i})|,
    \label{AELS}
\end{equation}
\begin{equation}
    \mathrm{RMSE} = \sqrt{\frac{1}{N}\sum_{i=1}^{N}(d_i - \hat{d_i})^2}.
    \label{RMSE}
\end{equation}

Threshold accuracy is a binary classification metric used in depth estimation tasks. It measures the probability of correct classification of depth estimation algorithms after binarizing the true depth values. We selected three thresholds of $1.25$, $1.25^2$ and $1.25^3$ for threshold accuracy calculation:
\begin{equation}
    \delta_{th}  = \frac{1}{N} \sum_{i=1}^{N} [\max(\frac{d_i}{\hat{d_i}}, \frac{\hat{d_i}}{d_i}) < \delta] ,\delta = 1.25, 1.25^2, 1.25^3,
    \label{Threshold accuracy}
\end{equation}
where $N$ is the number of samples, $d_i$ represents the ground truth depth, and $\hat{d_i}$ represents the predicted depth.

\subsection{Performance Results}
\label{subsec:Quantified Results}

\begin{table*}[t]
\vspace{0.3cm}
\centering
\resizebox{\linewidth}{!}{
	\begin{tabular}{l|ccc|ccc}
	\toprule
		Method &  $\delta<1.25 \uparrow$ & $\delta<1.25^{2} \uparrow$ & $\delta<1.25^{3} \uparrow$ & MARE $\downarrow$ & RMSE $\downarrow$ & AELS $\downarrow$ \\ \midrule
		 Baseline  &0.394 &0.683 &0.851 &0.388 &0.156 &1.167\\ 
          Baseline + Learnable Prompt  &0.429 &0.711 &0.878 &0.386 &0.145 &1.076\\ 
          Baseline + Learnable Prompt + Learnable Depth Bin  &\textbf{0.433} &0.715 &0.884 &0.378 &0.143 &1.066\\ \rowcolor{gray!20}
          \textbf{Baseline + Learnable Prompt + Learnable Depth Codebook}& 0.428 &\textbf{0.732} &\textbf{0.898} &\textbf{0.347} &\textbf{0.140} &\textbf{1.049}\\ 
	\bottomrule
	\end{tabular}
}
\vspace{-0.09cm}
\caption{\textbf{Impact of different algorithm components.} We report the performances with different design choices. `Learnable Depth Bin' means to learn a single class-independent depth bin. Our depth codebook is designed to learn a class-dependent depth bin for each scene.}
\vspace{-10pt}
\label{tab:ablation}
\end{table*}

\begin{table}[t]
% \vspace{0.3cm}
\centering
\resizebox{\linewidth}{!}{
	\begin{tabular}{l|ccc|ccc}
	\toprule
		Patch size &  $\delta<1.25 \uparrow$ & $\delta<1.25^{2} \uparrow$ & $\delta<1.25^{3} \uparrow$ & MARE $\downarrow$ & RMSE $\downarrow$ & AELS $\downarrow$ \\ \midrule
		 $8\times 8$  &0.336 &0.617 &0.806 &0.379 &0.179 &1.300\\ 
          $16\times 16$  &0.354 &0.630 &0.817 &0.514 &0.172 &1.229\\ 
          $32\times 32$  &\textbf{0.428} &\textbf{0.732} &\textbf{0.898} &\textbf{0.347} &\textbf{0.140} &\textbf{1.049}\\
	\bottomrule
	\end{tabular}
}
\vspace{-0.09cm}
\caption{\revise{\textbf{Effects of different patch sizes.} In our experiments, we set the patch size to $32\times 32$.}}
\vspace{-10pt}
\label{tab:ablation of patch size}
\end{table}

In Table \ref{tab:nyu_results}, we compare our results with other monocular depth estimation methods on NYU dataset. The upper half part is the prediction result of the supervised learning method, while the lower half part includes the performance results of the previous zero-shot depth estimation approach.
The results obtained by our method is presented in the last row. It is trained based on a few-shot setting
which randomly selects one picture per category. 
The lower bound row is obtained by making random predictions for each pixel within the depth ranging from 0 to 10 meters. 
Our few-shot-based method has notable improvement compared to the original zero-shot DepthCLIP\cite{zhang2022can}, exceeding the lower bound by a wider margin. And the adapted method has advantages in all metrics over other zero-shot transferring methods which have been pre-trained on specific datasets for monocular depth estimation (unsupervised KITTI video\cite{geiger2013vision}). Moreover, our method could obtain fairly performance of some fully-supervised methods like Make3D\cite{saxena2008make3d}, even surpassing it in some metrics. It has considerable accuracy under $\delta < 1.25^3$ and MARE deviation, along with notably lower RMSE of 1.049 (compared with 1.214 of Make3D\cite{saxena2008make3d} and 1.167 of DepthCLIP\cite{zhang2022can}). 
These extensive experimental results demonstrate the effectiveness of our few-shot approach in adapting the vision-language models for monocular depth estimation.

\revise{In Table \ref{tab:kitti_results}, we compare our results with other monocular depth estimation methods on KITTI dataset. Following the same principle of comparison of NYU dataset, our method still outperforms DepthCLIP by a substantial margin.}

\begin{figure}[!t]
\centering
\includegraphics[width=\linewidth]{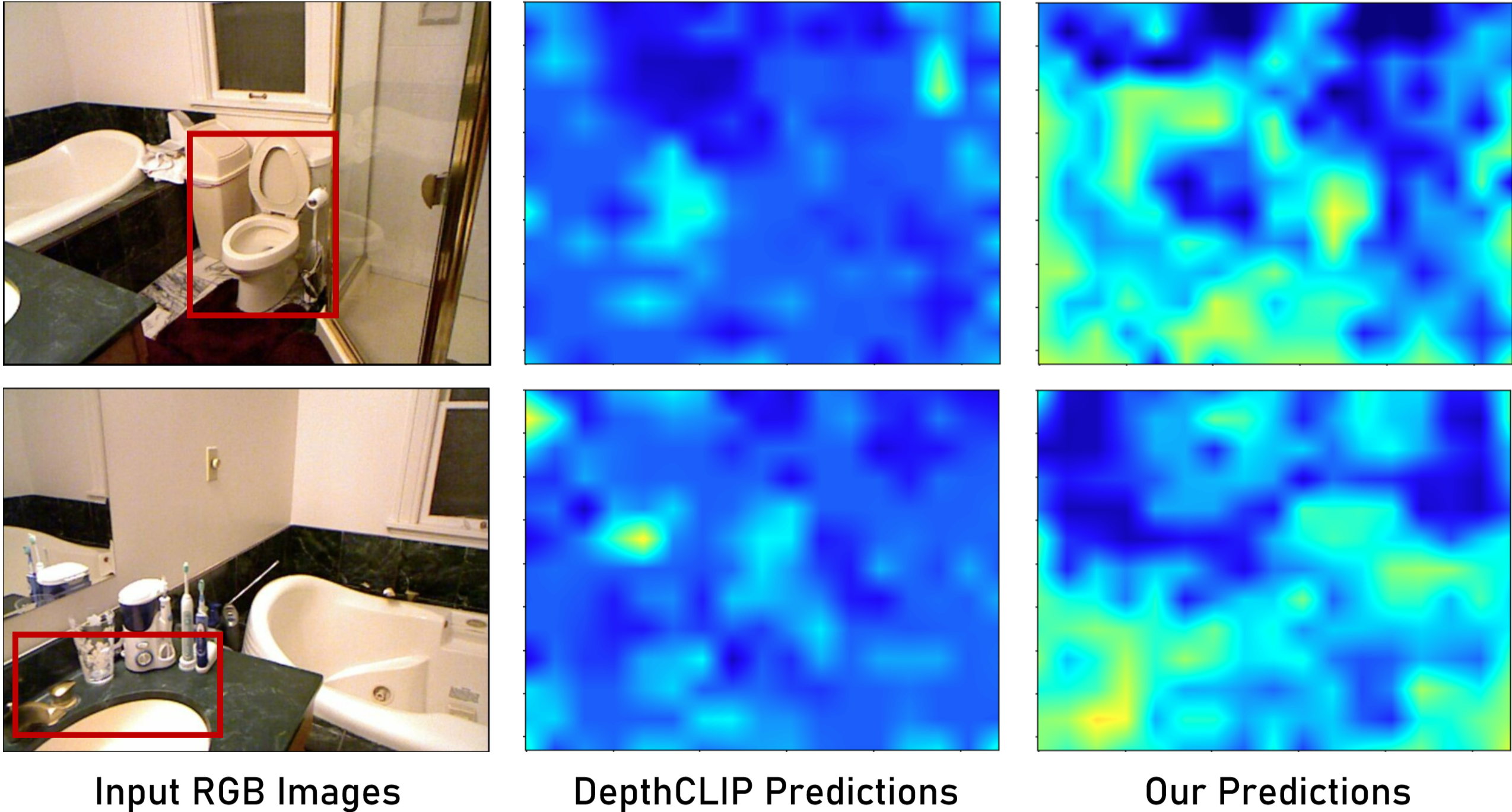}
% \vspace{-2pt}
\caption{\textbf{Qualitative comparisons on the NYU V2 dataset~\cite{silberman2012indoor}.} We present the depth estimation results obtained by DepthCLIP \cite{zhang2022can} and our proposed methods on three images to show the effectiveness of our proposed method.}
\vspace{-7pt}
\label{fig:qualitative}
\end{figure}

\subsection{Qualitative Comparisons}
In Figure \ref{fig:qualitative}, we present qualitative comparisons of our proposed method and DepthCLIP \cite{zhang2022can} method on the NYU V2 dataset~\cite{silberman2012indoor}. We include three images for performance comparisons. We can see that, our method leads to better depth predictions than DepthCLIP \cite{zhang2022can} in all three images, especially within the red rectangular regions. These results indicate that our approach effectively leverages few-shot training to adapt vision-language models, resulting in improved depth predictions.

\subsection{Ablation Studies}
\label{subsec:Ablation Studies}
To systematically evaluate our proposed method, we provide an empirical analysis of our design choices in this section. In Table \ref{tab:ablation}, we present the ablation study results. All the experiments are conducted on the NYU V2 dataset~\cite{silberman2012indoor}. 
Our method has two major new components, namely learnable prompt and learnable depth codebook introduced in Section \ref{sec:method}. In the first row of Table \ref{tab:ablation}, we report the results of the baseline method (DepthCLIP \cite{zhang2022can}). From the table, we can see that each algorithm component has a significant contribution to the overall performance.

\textbf{Learnable prompt design.}
In the previous vision-language models, the prompt is consisted of a grammatically correct English sentence, it helps fulfill the text input which was then passed through a text encoder to obtain the associated text features. 
In the context of the depth estimation task, the sentence ``This object is \texttt{[Depth CLASS]}" is utilized\cite{zhang2022can}. 
To optimize this, the words in the sentence are transformed into a vector that can be learned, and this vector is subsequently used to replace the input language sentence, forming a concatenated vector. We call this learnable prompts. 
The second row shows results with prompt tuning, it shows that with learnable prompt setting, our model could be better adapted to the training set. Note that, the addition of learnable prompts improves the despth estimation performance considerably.

\textbf{Learnable depth codebook design.} 
To enhance the model's generalization ability, our method introduces a class-dependent depth codebook. After obtaining the scene class by one-hot coding, we attach a specific learnable depth bin to each semantic language token. Then the selected bin would be combined with depth weights to obtain the final prediction. In other words, we have mapped the same semantic token to different class-dependent depth bin according to scene category. 
As we can see in Table \ref{tab:ablation}, the third row shows results with both learnable prompt and one learnable class-independent depth bin for all scenes. And the last row shows the final results, replacing the single depth bin with a scene-adapted depth codebook. we could notice the obvious improvement in all metrics except the percentage of $\delta < 1.25$. 
This shows the superiority of our learnable depth codebook.

\revise{
\textbf{Patch size.}
The patch size is related to the selected backbone, which is ResNet-50 in our experiments. Specifically, after feature extraction by CLIP's image encoder, the the obtained feature dimensions are the original size of the input image divided by 32. This is determined by the internal structure of ResNet-50 and the fixed parameters of the CLIP model. So in our experiments, we naturally set the patch size to 32. 
% The patch size cannot be increased to 64 because the pixel size of the image itself cannot be fully divided.
To ablate the effects of patch size, we conduct experiments with patch sizes 8 and 16 by using the middle layer features of the encoder as the extracted image features. The results are shown in Table \ref{tab:ablation of patch size}.
% To reduce the patch size, we can only use the middle layer features of the encoder as the extracted image features. Following this approach, we conducted ablation experiments with sizes set to 8 and 16, as shown in Table \ref{tab:ablation of patch size}. 
The results indicate that the patch size of 32 leads to the best performance.
% By the way, since the change in patch size does not involve training or updating new model parameters, and all calculations are matrix operations, so there will be no increase or decrease in training time.
}

\looseness=-1
\revise{
\textbf{Trade-of between efficiency and estimation accuracy.}
Fully-supervised methods demand a significant amount of training time to achieve satisfactory accuracy, whereas zero-shot approaches can obtain a basic level of accuracy without any training. Our method aims to strike a balance between training efficiency and generalization performance. Table \ref{tab:parameters} provides a comparative analysis of training time, parameter count, and RMSE across different methods, further illustrating this balance. For instance, DORN~\cite{fu2018deep} has 51M parameters and need to be trained by all taining data. The zero-shot DepthCLIP~\cite{zhang2022can}  incurs no training cost in terms of time and parameters but yields the highest RMSE.  In contrast, our approach only updates 8K parameters by training on a mere 27 images for 70 minutes, and yet achieves a lower RMSE.
}

% Our method tries to find a balance between training cost and generalization capabilities, which lies in the improvement in accuracy brought about by minimal training data. We use Table \ref{tab:parameters} to further explain the training cost by the number of parameters which need to be updated during training. According to DORN~\cite{fu2018deep}, its totally new designed encoder has 51M parameters waiting to be trained by all the data in training set. This supervised pattern spends more time to get a great performance. The zero-shot DepthCLIP~\cite{zhang2022can} which is free of training cost nothing in time and parameters gain the poorest RMSE. However, our method updates only 8k parameters by training 27 images but still increases the RMSE result.

\begin{table}[t]
\centering
\resizebox{\linewidth}{!}{
	\begin{tabular}{l|ccccc}
	\toprule
		Method & Supervision & Time &   Params. & RMSE $\downarrow$\\ \midrule
            DORN~\cite{fu2018deep}& full&  $>$ 30 h  & 51M  & 0.509\\
            DepthCLIP~\cite{zhang2022can}& 0-shot & 0 & 0   &  1.147\\
            Ours & 1-shot & 70 min & 8K   &  1.049\\
  
	\bottomrule
	\end{tabular}
}
\vspace{-0.05cm}
\caption{\revise{\textbf{Trade-of between efficiency and estimation accuracy.} We report the training time and number of parameters to be updated of fully-supervised, zero-shot, and our few-shot methods. We conduct the experiment on one single RTX 3090 GPU.}}
\label{tab:parameters}
\vspace{-10pt}
\end{table}

\section{Conclusion}
\label{sec:conclusion}
In this paper, we propose a few-shot-based method that aims to adapt CLIP for monocular depth estimation. This approach focuses on striking a balance between training costs and generalization ability. Specifically, it assigns varying depth bin to different scenes, allowing the model to select the appropriate bins during inference. Furthermore, we introduce a learnable prompt to preprocess the input text, facilitating the conversion of easily understandable human text into vectors that are readily interpreted by the model.
Remarkably, with only one image per scene for training, our extensive experiments on the NYU V2 and KITTI dataset demonstrate that our method surpasses the previous state-of-the-art approach by up to 10.6\%.

%%%%%%%%% REFERENCES
\balance
{\small
\bibliographystyle{ieee_fullname}
\bibliography{egbib}

\begin{thebibliography}{10}\itemsep=-1pt

\bibitem{alayrac2022flamingo}
Jean-Baptiste Alayrac, Jeff Donahue, Pauline Luc, Antoine Miech, Iain Barr,
  Yana Hasson, Karel Lenc, Arthur Mensch, Katherine Millican, Malcolm Reynolds,
  et~al.
\newblock Flamingo: a visual language model for few-shot learning.
\newblock In {\em NeurIPS}, volume~35, pages 23716--23736, 2022.

\bibitem{brown2020language}
Tom Brown, Benjamin Mann, Nick Ryder, Melanie Subbiah, Jared~D Kaplan, Prafulla
  Dhariwal, Arvind Neelakantan, Pranav Shyam, Girish Sastry, Amanda Askell,
  et~al.
\newblock Language models are few-shot learners.
\newblock In {\em NeurIPS}, volume~33, pages 1877--1901, 2020.

\bibitem{casser2019depth}
Vincent Casser, Soeren Pirk, Reza Mahjourian, and Anelia Angelova.
\newblock Depth prediction without the sensors: Leveraging structure for
  unsupervised learning from monocular videos.
\newblock In {\em AAAI}, volume~33, pages 8001--8008, 2019.

\bibitem{chang2021transformer}
Wenjie Chang, Yueyi Zhang, and Zhiwei Xiong.
\newblock Transformer-based monocular depth estimation with attention
  supervision.
\newblock In {\em BMVC}, 2021.

\bibitem{chen2023plot}
Guangyi Chen, Weiran Yao, Xiangchen Song, Xinyue Li, Yongming Rao, and Kun
  Zhang.
\newblock Prompt learning with optimal transport for vision-language models.
\newblock In {\em ICLR}, 2023.

\bibitem{deng2009imagenet}
Jia Deng, Wei Dong, Richard Socher, Li-Jia Li, Kai Li, and Li Fei-Fei.
\newblock Imagenet: A large-scale hierarchical image database.
\newblock In {\em CVPR}, pages 248--255, 2009.

\bibitem{desai2021virtex}
Karan Desai and Justin Johnson.
\newblock Virtex: Learning visual representations from textual annotations.
\newblock In {\em CVPR}, pages 11162--11173, 2021.

\bibitem{duan2022multi}
Jiali Duan, Liqun Chen, Son Tran, Jinyu Yang, Yi Xu, Belinda Zeng, and Trishul
  Chilimbi.
\newblock Multi-modal alignment using representation codebook.
\newblock In {\em CVPR}, pages 15651--15660, 2022.

\bibitem{fang2016fast}
Shuai Fang, Ren Jin, and Yang Cao.
\newblock Fast depth estimation from single image using structured forest.
\newblock In {\em ICIP}, pages 4022--4026. IEEE, 2016.

\bibitem{fu2018deep}
Huan Fu, Mingming Gong, Chaohui Wang, Kayhan Batmanghelich, and Dacheng Tao.
\newblock Deep ordinal regression network for monocular depth estimation.
\newblock In {\em CVPR}, pages 2002--2011, 2018.

\bibitem{gao2021clip}
Peng Gao, Shijie Geng, Renrui Zhang, Teli Ma, Rongyao Fang, Yongfeng Zhang,
  Hongsheng Li, and Yu Qiao.
\newblock Clip-adapter: Better vision-language models with feature adapters.
\newblock {\em IJCV}, pages 1--15, 2023.

\bibitem{geiger2013vision}
Andreas Geiger, Philip Lenz, Christoph Stiller, and Raquel Urtasun.
\newblock Vision meets robotics: The kitti dataset.
\newblock {\em The International Journal of Robotics Research},
  32(11):1231--1237, 2013.

\bibitem{gomez2017self}
Lluis Gomez, Yash Patel, Mar{\c{c}}al Rusinol, Dimosthenis Karatzas, and CV
  Jawahar.
\newblock Self-supervised learning of visual features through embedding images
  into text topic spaces.
\newblock In {\em CVPR}, pages 4230--4239, 2017.

\bibitem{huang2022idea}
Xinyu Huang, Youcai Zhang, Ying Cheng, Weiwei Tian, Ruiwei Zhao, Rui Feng,
  Yuejie Zhang, Yaqian Li, Yandong Guo, and Xiaobo Zhang.
\newblock Idea: Increasing text diversity via online multi-label recognition
  for vision-language pre-training.
\newblock In {\em ACM MM}, pages 4573--4583, 2022.

\bibitem{jackson2019style}
Philip~TG Jackson, Amir~Atapour Abarghouei, Stephen Bonner, Toby~P Breckon, and
  Boguslaw Obara.
\newblock Style augmentation: data augmentation via style randomization.
\newblock In {\em CVPRW}, volume~6, pages 10--11, 2019.

\bibitem{jia2021scaling}
Chao Jia, Yinfei Yang, Ye Xia, Yi-Ting Chen, Zarana Parekh, Hieu Pham, Quoc Le,
  Yun-Hsuan Sung, Zhen Li, and Tom Duerig.
\newblock Scaling up visual and vision-language representation learning with
  noisy text supervision.
\newblock In {\em ICML}, pages 4904--4916, 2021.

\bibitem{jia2022vpt}
Menglin Jia, Luming Tang, Bor-Chun Chen, Claire Cardie, Serge Belongie, Bharath
  Hariharan, and Ser-Nam Lim.
\newblock Visual prompt tuning.
\newblock In {\em ECCV}, 2022.

\bibitem{la2022adversarial}
Riccardo La~Grassa, Ignazio Gallo, Cristina Re, Gabriele Cremonese, Nicola
  Landro, Claudio Pernechele, Emanuele Simioni, and Mattia Gatti.
\newblock An adversarial generative network designed for high-resolution
  monocular depth estimation from 2d hirise images of mars.
\newblock {\em Remote Sensing}, 14(18):4619, 2022.

\bibitem{lei2021less}
Jie Lei, Linjie Li, Luowei Zhou, Zhe Gan, Tamara~L Berg, Mohit Bansal, and
  Jingjing Liu.
\newblock Less is more: Clipbert for video-and-language learning via sparse
  sampling.
\newblock In {\em CVPR}, pages 7331--7341, 2021.

\bibitem{lei2015predicting}
Jimmy Lei~Ba, Kevin Swersky, Sanja Fidler, et~al.
\newblock Predicting deep zero-shot convolutional neural networks using textual
  descriptions.
\newblock In {\em CVPR}, pages 4247--4255, 2015.

\bibitem{lester2021power}
Brian Lester, Rami Al-Rfou, and Noah Constant.
\newblock The power of scale for parameter-efficient prompt tuning.
\newblock In {\em EMNLP}, pages 3045--3059, 2021.

\bibitem{li2021grounded}
Liunian~Harold Li, Pengchuan Zhang, Haotian Zhang, Jianwei Yang, Chunyuan Li,
  Yiwu Zhong, Lijuan Wang, Lu Yuan, Lei Zhang, Jenq-Neng Hwang, et~al.
\newblock Grounded language-image pre-training.
\newblock In {\em CVPR}, pages 10965--10975, 2022.

\bibitem{li2019few}
Shuai Li, Jiaying Shi, Wenfeng Song, Aimin Hao, and Hong Qin.
\newblock Few-shot learning for monocular depth estimation based on local
  object relationship.
\newblock In {\em ICTAI}, pages 1221--1228. IEEE, 2019.

\bibitem{li2021prefix}
Xiang~Lisa Li and Percy Liang.
\newblock Prefix-tuning: Optimizing continuous prompts for generation.
\newblock In {\em ACL}, pages 4582--4597, 2021.

\bibitem{li2022depthformer}
Zhenyu Li, Zehui Chen, Xianming Liu, and Junjun Jiang.
\newblock Depthformer: Exploiting long-range correlation and local information
  for accurate monocular depth estimation.
\newblock {\em Machine Intelligence Research}, pages 1--18, 2023.

\bibitem{lu2019vilbert}
Jiasen Lu, Dhruv Batra, Devi Parikh, and Stefan Lee.
\newblock Vilbert: Pretraining task-agnostic visiolinguistic representations
  for vision-and-language tasks.
\newblock In {\em NeurIPS}, volume~32, 2019.

\bibitem{lu2022prompt}
Yuning Lu, Jianzhuang Liu, Yonggang Zhang, Yajing Liu, and Xinmei Tian.
\newblock Prompt distribution learning.
\newblock In {\em CVPR}, pages 5206--5215, 2022.

\bibitem{mahjourian2018unsupervised}
Reza Mahjourian, Martin Wicke, and Anelia Angelova.
\newblock Unsupervised learning of depth and ego-motion from monocular video
  using 3d geometric constraints.
\newblock In {\em CVPR}, pages 5667--5675, 2018.

\bibitem{mel2022end}
Mazen Mel, Muhammad Siddiqui, and Pietro Zanuttigh.
\newblock End-to-end learning for joint depth and image reconstruction from
  diffracted rotation.
\newblock {\em arXiv preprint arXiv:2204.07076}, 2022.

\bibitem{ming2021deep}
Yue Ming, Xuyang Meng, Chunxiao Fan, and Hui Yu.
\newblock Deep learning for monocular depth estimation: A review.
\newblock {\em Neurocomputing}, 438:14--33, 2021.

\bibitem{radford2021learning}
Alec Radford, Jong~Wook Kim, Chris Hallacy, Aditya Ramesh, Gabriel Goh,
  Sandhini Agarwal, Girish Sastry, Amanda Askell, Pamela Mishkin, Jack Clark,
  et~al.
\newblock Learning transferable visual models from natural language
  supervision.
\newblock In {\em ICML}, 2021.

\bibitem{radford2019language}
Alec Radford, Jeffrey Wu, Rewon Child, David Luan, Dario Amodei, Ilya
  Sutskever, et~al.
\newblock Language models are unsupervised multitask learners.
\newblock {\em OpenAI Blog}, 1(8):9, 2019.

\bibitem{roy2016monocular}
Anirban Roy and Sinisa Todorovic.
\newblock Monocular depth estimation using neural regression forest.
\newblock In {\em CVPR}, pages 5506--5514, 2016.

\bibitem{sariyildiz2020learning}
Mert~Bulent Sariyildiz, Julien Perez, and Diane Larlus.
\newblock Learning visual representations with caption annotations.
\newblock In {\em ECCV}, pages 153--170, 2020.

\bibitem{saxena2008make3d}
Ashutosh Saxena, Min Sun, and Andrew~Y Ng.
\newblock Make3d: Learning 3d scene structure from a single still image.
\newblock {\em IEEE TPAMI}, 31(5):824--840, 2008.

\bibitem{schick2020exploiting}
Timo Schick and Hinrich Sch{\"u}tze.
\newblock Exploiting cloze questions for few shot text classification and
  natural language inference.
\newblock In {\em EACL}, pages 255--269, 2021.

\bibitem{schick2020s}
Timo Schick and Hinrich Sch{\"u}tze.
\newblock It's not just size that matters: Small language models are also
  few-shot learners.
\newblock In {\em NAACL}, pages 2339--2352, 2021.

\bibitem{silberman2012indoor}
Nathan Silberman, Derek Hoiem, Pushmeet Kohli, and Rob Fergus.
\newblock Indoor segmentation and support inference from rgbd images.
\newblock {\em ECCV}, 7576:746--760, 2012.

\bibitem{tang2023cross}
Yushun Tang, Qinghai Guo, and Zhihai He.
\newblock Cross-inferential networks for source-free unsupervised domain
  adaptation.
\newblock In {\em ICIP}, pages 96--100. IEEE, 2023.

\bibitem{tang2023neuro}
Yushun Tang, Ce Zhang, Heng Xu, Shuoshuo Chen, Jie Cheng, Luziwei Leng, Qinghai
  Guo, and Zhihai He.
\newblock Neuro-modulated hebbian learning for fully test-time adaptation.
\newblock In {\em CVPR}, pages 3728--3738, 2023.

\bibitem{tosi2019learning}
Fabio Tosi, Filippo Aleotti, Matteo Poggi, and Stefano Mattoccia.
\newblock Learning monocular depth estimation infusing traditional stereo
  knowledge.
\newblock In {\em CVPR}, pages 9799--9809, 2019.

\bibitem{wang2022learning}
Feng Wang, Manling Li, Xudong Lin, Hairong Lv, Alexander~G Schwing, and Heng
  Ji.
\newblock Learning to decompose visual features with latent textual prompts.
\newblock In {\em ICLR}, 2023.

\bibitem{wang2022efficient}
Ning Wang, Jiangrong Xie, Hang Luo, Qinglin Cheng, Jihao Wu, Mingbo Jia, and
  Linlin Li.
\newblock Efficient image captioning for edge devices.
\newblock In {\em AAAI}, volume~37, pages 2608--2616, 2023.

\bibitem{yao2021cpt}
Yuan Yao, Ao Zhang, Zhengyan Zhang, Zhiyuan Liu, Tat-Seng Chua, and Maosong
  Sun.
\newblock Cpt: Colorful prompt tuning for pre-trained vision-language models.
\newblock {\em arXiv preprint arXiv:2109.11797}, 2021.

\bibitem{yu2022coca}
Jiahui Yu, Zirui Wang, Vijay Vasudevan, Legg Yeung, Mojtaba Seyedhosseini, and
  Yonghui Wu.
\newblock Coca: Contrastive captioners are image-text foundation models.
\newblock {\em TMLR}, 2022.

\bibitem{zhang2020unsupervised}
Mingliang Zhang, Xinchen Ye, Xin Fan, and Wei Zhong.
\newblock Unsupervised depth estimation from monocular videos with hybrid
  geometric-refined loss and contextual attention.
\newblock {\em Neurocomputing}, 379:250--261, 2020.

\bibitem{zhang2021vt}
Renrui Zhang, Longtian Qiu, Wei Zhang, and Ziyao Zeng.
\newblock Vt-clip: Enhancing vision-language models with visual-guided texts.
\newblock {\em arXiv preprint arXiv:2112.02399}, 2021.

\bibitem{zhang2022can}
Renrui Zhang, Ziyao Zeng, Ziyu Guo, and Yafeng Li.
\newblock Can language understand depth?
\newblock In {\em ACM MM}, pages 6868--6874, 2022.

\bibitem{zhang2023unsupervised}
Yi Zhang, Ce Zhang, Xueting Hu, and Zhihai He.
\newblock Unsupervised prototype adapter for vision-language models.
\newblock In {\em PRCV}, 2023.

\bibitem{zhang2023bdc}
Yi Zhang, Ce Zhang, Zihan Liao, Yushun Tang, and Zhihai He.
\newblock Bdc-adapter: Brownian distance covariance for better vision-language
  reasoning.
\newblock In {\em BMVC}, 2023.

\bibitem{zhang2023cross}
Yi Zhang, Ce Zhang, Yushun Tang, and Zhihai He.
\newblock Cross-modal concept learning and inference for vision-language
  models.
\newblock {\em arXiv preprint arXiv:2307.15460}, 2023.

\bibitem{zhou2022conditional}
Kaiyang Zhou, Jingkang Yang, Chen~Change Loy, and Ziwei Liu.
\newblock Conditional prompt learning for vision-language models.
\newblock In {\em CVPR}, pages 16816--16825, 2022.

\bibitem{zhou2022learning}
Kaiyang Zhou, Jingkang Yang, Chen~Change Loy, and Ziwei Liu.
\newblock Learning to prompt for vision-language models.
\newblock {\em IJCV}, 130(9):2337--2348, 2022.

\bibitem{zhou2022unsupervised}
Mingyang Zhou, Licheng Yu, Amanpreet Singh, Mengjiao Wang, Zhou Yu, and Ning
  Zhang.
\newblock Unsupervised vision-and-language pre-training via retrieval-based
  multi-granular alignment.
\newblock In {\em CVPR}, pages 16485--16494, 2022.

\end{thebibliography}
}

\end{document}